\documentclass[letterpaper, 10 pt, conference]{ieeeconf}  

\IEEEoverridecommandlockouts                              

\usepackage{graphics} 
\usepackage{epsfig} 
\usepackage{mathptmx} 
\usepackage{times} 
\usepackage{amsmath} 
\usepackage{amssymb}  
\usepackage{wasysym}
\usepackage[dvipsnames]{xcolor}
\usepackage[noadjust]{cite}
\usepackage{tabularx}
\usepackage{url}
\usepackage{xspace}
\usepackage{gensymb}
\usepackage{booktabs}
\usepackage{caption} 
\let\labelindent\relax
\usepackage{enumitem}
\usepackage{soul}
\newcommand{\figref}[1]{Fig.~\ref{#1}}     
\newcommand{\tableref}[1]{Table~\ref{#1}}
\newcommand{\secref}[1]{Section~\ref{#1}}

\newcommand{\rel}[1]{\textcolor{black}{\xspace#1}}

\title{\LARGE \bf
In-the-Wild Compliant Manipulation with UMI-FT
}

\author{Hojung Choi*$^{1}$, Yifan Hou*$^{1}$, Chuer Pan$^{1}$, Seongheon Hong$^{2}$, Austin Patel$^{1}$, Xiaomeng Xu$^{1}$, \\ Mark R. Cutkosky$^{2}$, and Shuran Song$^{1}$%
\thanks{*Equal contribution.}%
\thanks{$^1$Department of Electrical Engineering, Stanford University, CA, USA.}%
\thanks{$^2$Department of Mechanical Engineering, Stanford University, CA, USA.}%
}

\IEEEoverridecommandlockouts 

\IEEEaftertitletext{
    \begin{center}
        \vspace{-10pt} 
        \includegraphics[width=\textwidth]{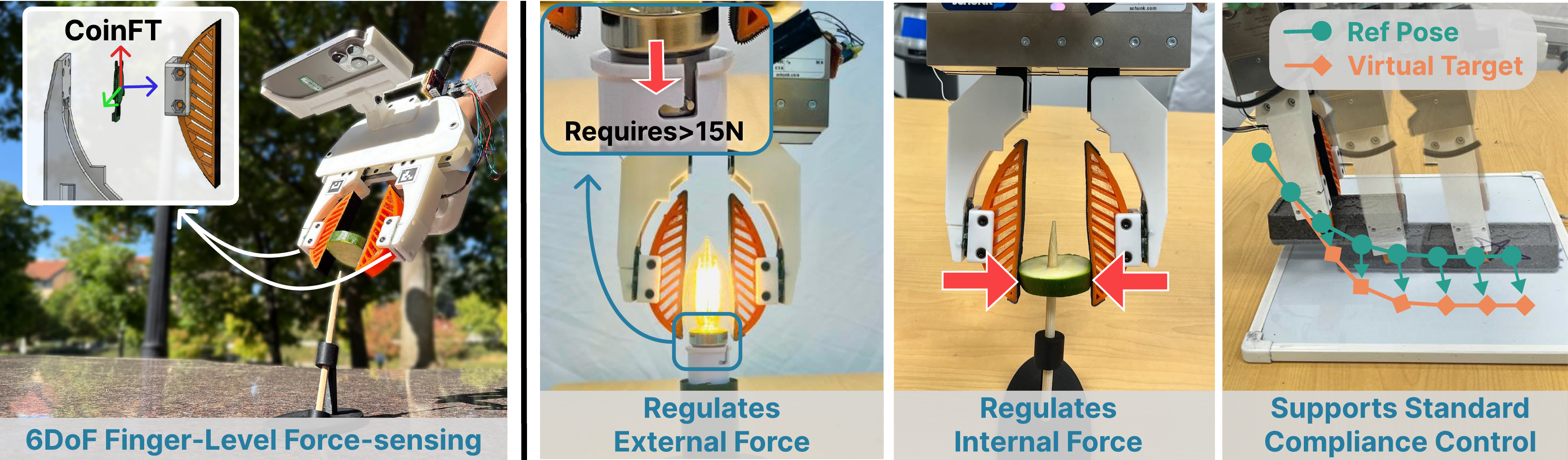}
        \captionof{figure}{{\label{fig:teaser} \textbf{UMI-FT design and capabilities.} Left: UMI-FT uses coin-sized force/torque sensors to record the wrench at each compliant fingertip. Right: we evaluated UMI-FT in three tasks: lightbulb insertion, skewering zucchini, and whiteboard wiping. In each task, finger-level force sensing captures external contact forces and internal grasp forces, similar to forces experienced by a human operator during demonstrations. We use the external force measurement in a standard compliance controller to track a virtual spring target.
        }}
        \vspace{2pt}
    \end{center}
}

\begin{document}
\maketitle

\thispagestyle{empty}
\pagestyle{empty}

\begin{abstract}

Many manipulation tasks require careful force modulation. With insufficient force the task may fail, while excessive force could cause damage. The high cost, bulky size and fragility of commercial force/torque (F/T) sensors have limited large-scale, force-aware policy learning. We introduce UMI-FT, a handheld data-collection platform that mounts compact, six-axis force/torque sensors 
on each finger, enabling finger-level wrench measurements alongside RGB, depth, and pose. Using the multimodal data collected from this device, we train an adaptive compliance policy
that predicts position targets, grasp force, and stiffness for execution on standard compliance controllers.
In evaluations on three contact-rich, force-sensitive tasks (whiteboard wiping, skewering zucchini, and lightbulb insertion), UMI-FT enables policies that reliably regulate external contact forces and internal grasp forces, outperforming baselines that lack compliance or force sensing. UMI-FT offers a scalable path to learning compliant manipulation from in-the-wild demonstrations. We open-source the hardware and software to facilitate broader adoption at: {\color{RoyalBlue}\url{https://umi-ft.github.io/}.}









\end{abstract}


\section{Introduction}



Handheld manipulation devices, such as UMI~\cite{chi2024universal}, are becoming increasingly popular for manipulation data collection \cite{bhirangi2024anyskinplugandplayskinsensing, liu2024maniwav, zhu2025touch} due to their flexibility and natural haptic feedback. However, most existing handheld data collection systems focus only on position and visual information, without considering force sensing. Yet force sensing is a key element of human manipulation, enabling careful contact modulation that is especially critical when handling soft or fragile objects.

In this paper, we introduce UMI-FT, a novel force-sensing handheld data collection device. Our key improvement over UMI is the integration of CoinFT \cite{choi2025coinftcoinsizedcapacitive6axis}, a custom, compact force–torque sensor mounted at each finger. This design captures the complete force information present during a demonstration (both the external forces transmitted through the device structure and the internal grasp forces exerted by the user), thereby providing an authentic representation of what the human demonstrator experiences.

Prior approaches to collecting force data can be broadly divided into two categories. The first employs a wrist-mounted commercial six-axis F/T sensor (e.g., \cite{galbally2022elly, liu2025forcemimic}). In contrast, UMI-FT offers two key advantages:

\begin{itemize}[leftmargin=4mm]
    \item \textbf{Finger-level force}. The compact design of CoinFT enables a sensor to be mounted at each finger. This placement bypasses the gripper mass, providing more accurate and low-latency contact force measurements, and allows the robot to regulate internal grasp forces.
    \item \textbf{Scalability}. CoinFT is low-cost (\$10 BOM per sensor) and resilient to impact, making it more scalable than expensive, fragile commercial F/T sensors.
\end{itemize}

The second category of prior sensing solutions is to install \textit{tactile sensors} on fingers \cite{liu2024maniwav, zhu2025touch, bhirangi2024anyskinplugandplayskinsensing}. Here again, UMI-FT provides the following key advantages:
\begin{itemize}[leftmargin=4mm]
    \item \textbf{Standard and consistent measurements}. The measurements from UMI-FT can be directly calibrated using a commercial F/T sensor, providing consistent data across different units. Consistency is maintained regardless of installation location, since measured wrenches can be transformed into any desired coordinate frame (typically the robot tool frame).
    \item \textbf{Compatibility with standard compliance control}. The wrench measurements from UMI-FT can be used with standard compliance controllers (e.g., admittance or impedance control) to implement a stable, accurate, and user-specified compliance profile.
    \item \textbf{Full finger sensing}. Unlike tactile sensors which must physically cover all potential contact regions, UMI-FT can measure contact forces anywhere along the finger body, as long as the load path passes through the sensor.
\end{itemize}

To showcase these advantages, we train an Adaptive Compliance Policy (ACP) \cite{hou2025adaptivecompliancepolicylearning} using data collected through UMI-FT, and demonstrate safe, yet forceful interactions. Ablation studies highlight the benefits of explicit finger-level force data, particularly for tasks requiring forceful manipulation, such as wiping a whiteboard, skewering a zucchini and lightbulb insertion.

In summary, our main contributions are as follows.
\begin{itemize}[leftmargin=4mm]
    \item \textbf{UMI with per-finger force/torque sensing.} A portable and scalable handheld device equipped with per-finger F/T sensors, capturing both external contact forces and internal grasp force that human demonstrators naturally exploit through haptic feedback.

    \item \textbf{A control architecture for compliant manipulation}. We augment ACP with the ability to regulate both external contact forces and internal grasp force, enabling robust and safe manipulation without requiring bulky, expensive F/T sensors. 
    
    \item \textbf{Experimental validation.} We demonstrate significant improvements from both finger-level force sensing and robot compliance enabled by UMI-FT, across tasks that require safe yet forceful manipulation.

    \item \textbf{Open-source hardware and software}, to facilitate large-scale adoption and encourage broader research on multimodal robot data collection.
    
    
\end{itemize}

\section{Related Work}
\label{sec:related-work}
\subsection{Force/Tactile Sensing for Manipulation Data Collection and Policy Learning}
Prior imitation learning systems have attempted to obtain direct force/torque measurements through motor joint current~\cite{wu2025robocopilot} or wrist-mounted commercial force/torque (F/T) sensors~\cite{liu2025forcemimic,hou2025adaptivecompliancepolicylearning}. As explained in the Introduction, these systems lack finger-level sensing and are hard to scale due to their size, fragility, and cost.

Tactile sensors have compact sizes and provide rich contact information directly from the fingertip.
A wide range of tactile sensors have been developed to capture different tactile features, including contact pressure and location, multi-axial force, slippage, temperature, and proximity~\cite{li2020review,luo2025tactile}.
Vision-based tactile sensors have also grown in popularity due to their high resolution and sensitivity~\cite{li2024vision, yuan2017gelsight}. Tactile sensors have been integrated into a variety of data collection interfaces, including teleoperation~\cite{huang20243d,zhao2025polytouch}, handheld devices~\cite{zhu2025touch,liu2024maniwav}, wearable systems~\cite{xu2025dexumiusinghumanhand}, and kinesthetic teaching~\cite{hou2025adaptivecompliancepolicylearning,DexForce_Chen2025}.  

These modalities have been shown to benefit policy learning at multiple levels. Contact microphones can capture dynamic events and material properties~\cite{liu2024maniwav}. High-resolution tactile arrays and vision-based tactile sensors can infer object pose and contact intensity~\cite{huang20243d,zhu2025touch,zhao2025polytouch,li2022see, liu2023enhancing}. Wearable systems provide demonstrators with direct haptic feedback, embedding fine-grained strategies that can transfer to robot policies~\cite{xu2025dexumiusinghumanhand}. Direct F/T measurements through kinesthetic interfaces have likewise proven effective for learning compliant behaviors, both at the wrist~\cite{hou2025adaptivecompliancepolicylearning} and fingertip level~\cite{DexForce_Chen2025}.  

Our objective is to approximate what a human user experiences while operating a hand-held gripper device. Primarily, this includes dynamic forces, transmitted through the structure of the device and a sense of the applied grasp force. For this reason, miniature force/torque sensors at the fingertips 
(e.g. as in \cite{el2024compact}) are particularly of interest. In particular, we choose CoinFT \cite{choi2025coinftcoinsizedcapacitive6axis} for its small size (20\,mm\diameter $\times$3\,mm thick) and low weight, robustness, and low cost---allowing replacement should damage occur.



\subsection{Compliance and Compliance Control}
Compliance refers to the elastic behavior of a physical body under external force, commonly described by a combination of stiffness, inertia and damping~\cite{villani2016force}. At low speed, only stiffness remains significant. A suitable stiffness profile can improve the disturbance rejection capability of manipulation systems~\cite{8794366,hou2020manipulation}. Compliance is often necessary for stability in contact-rich manipulations~\cite{lee2019making,li2022see,luo2025precise, xu2025compliant}. 

When force feedback is available, specific compliance profiles can be achieved with standard impedance~\cite{hogan1985impedance} control on backdrivable robots, or admittance control~\cite{maples1986experiments} on stiff, high accuracy robots.

\section{Multimodal Sensing using UMI-FT}
\label{sec:methods-HW}

In this section, we introduce our hardware suite for collecting multimodal data from human demonstrations.

\subsection{UMI-FT Design}

UMI-FT is a modified version of the original UMI device, designed to collect multimodal sensory data including vision, pose, and six-axis F/T signals at each finger, during human demonstrations (\figref{fig:teaser}). The system integrates CoinFT sensors mounted at both fingers and an iPhone\,15\,Pro. The iPhone is rigidly attached to the UMI-FT structure with a 15\degree \, tilt to better capture the finger workspace and ArUco markers (for measuring gripper width). It provides synchronized main RGB (approximately 80\degree \,diagonal FoV), ultrawide RGB (120\degree \,FoV), depth, and pose data via ARKit. The main RGB and pose stream are recorded at 60\,Hz, ultrawide RGB at 10\,Hz, and depth at 30\,Hz.

Each finger is equipped with a CoinFT sensor, which measures wrenches at 360\,Hz. Its unit weight, 2\,g, has minimal impact on overall weight of the UMI-FT device. These sensors are connected to a USB-serial bus via I\textsuperscript{2}C, and the bus is connected to a laptop with a USB cable for data logging. CoinFT data is timestamped using internet time and synchronized with the iPhone data streams during post-processing.

Significant modifications were made to the original fin-ray finger design of the UMI to accommodate the constraints of CoinFT (\figref{fig:teaser}). While CoinFT is mechanically robust under compressive loading, its tensile tolerance is lower. During manipulation, excessive moments caused by grasp forces and long moment arms can lead to delamination of the sensor. To mitigate this, each CoinFT is mounted closer to the fingertip, reducing the moment arm and ensuring that grasping primarily results in compressive forces. This placement allows the UMI-FT fingers to perform a stronger grasp while preserving mechanical integrity. 

The CoinFT itself was also redesigned to enhance robustness and increase sensing range, with a tradeoff in sensitivity. The dielectric layer consists of pillars with oval cross section, where in the outermost layer the major and minor axes are approximately 4\,mm and 0.5\,mm, respectively, in contrast to circular pillars with a 0.2\,mm diameter from the original design \cite{choi2025coinftcoinsizedcapacitive6axis}.

\subsection{UMI-FT Finger Calibration}

Unlike common commercial F/T sensors such as the Gamma (ATI), CoinFT exhibits a nonlinear signal profile due to both its mechanical properties and the physics of capacitive sensing \cite{choi2025coinftcoinsizedcapacitive6axis}. The signal behavior cannot be easily modeled using traditional linear techniques under large loads that may occur during manipulation.



\begin{figure}[tp]
\centering
	\vspace{1.5mm}
	\includegraphics[width=\linewidth]{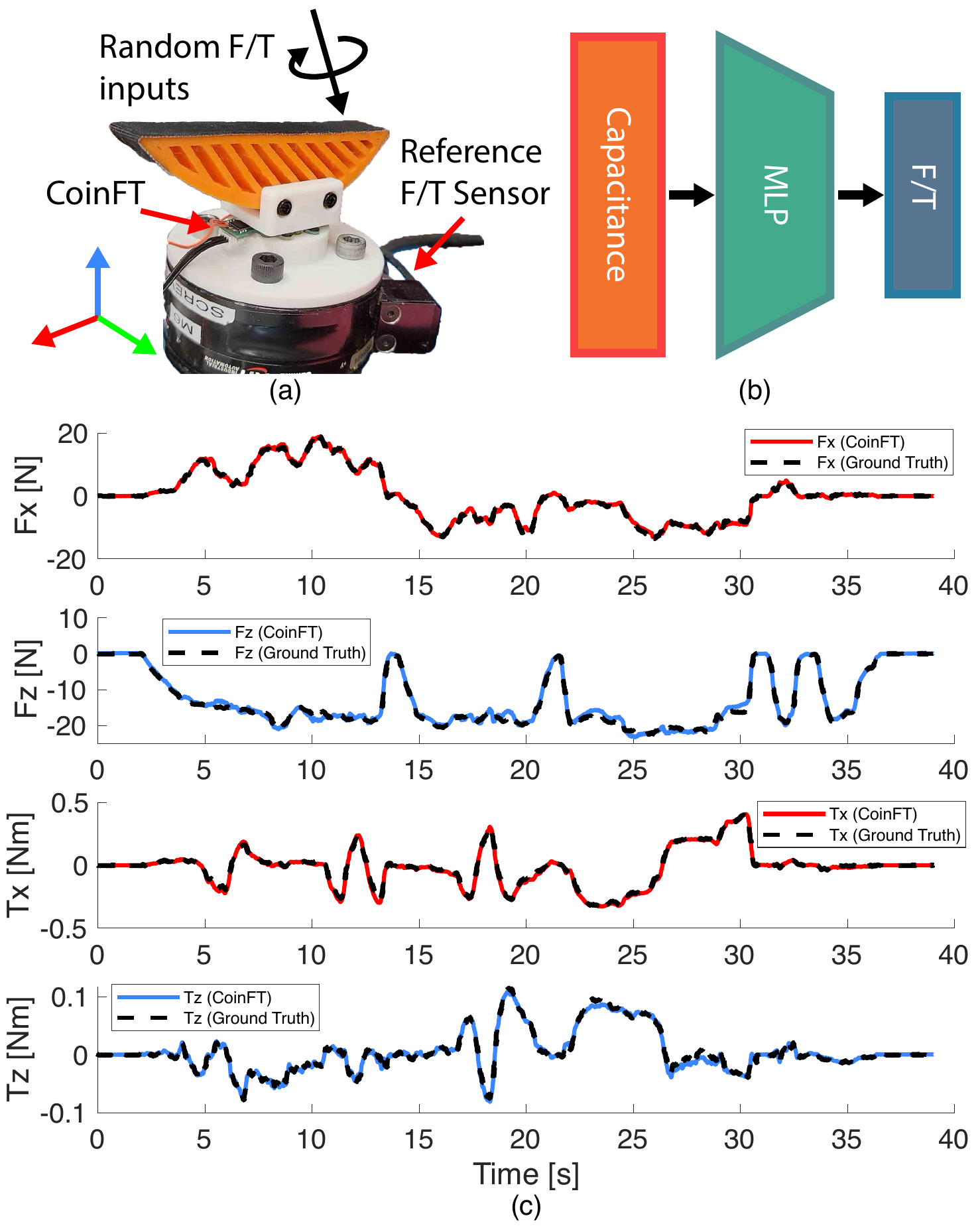}
	\caption{UMI-FT finger calibration. (a) Raw capacitance data from CoinFT and the corresponding F/T from a reference sensor (Gamma, ATI) are collected with random F/T inputs. (b) The raw capacitance is mapped to F/T through an MLP layer. (c) Calibration results on unseen input. Only one of the shear axes (x, y) were plotted due to similarity. }
	\label{fig:calibration}
	\vspace{-7pt}
\end{figure}

To achieve accurate sensing, we calibrate each CoinFT sensor in situ, with the UMI-FT fingertip mounted, ensuring that the calibration model captures the effects of the full finger structure (\figref{fig:calibration}\,(a)). The Gamma (ATI) is used as ground truth reference. During data collection, a human operator applies randomized combinations of forces and torques along the UMI-FT finger, covering a target range of 25\,N in the normal direction, 20\,N in shear, and up to 500\,mNm in moment.

The mapping from capacitance to force/torque is learned using a multi-layer perceptron (MLP) with five hidden fully connected layers, with layer widths of 128, 64, 36, 24, and 12, respectively (\figref{fig:calibration}\,(b)). Each hidden layer employs ReLU activation, and the final layer performs a six-dimensional regression to estimate the full six-axis F/T. This model achieves a mean squared error of 0.18\,N, 0.15\,N, 0.58\,N, 159\,mNm, 231\,mNm, 17\,mNm for $Fx$, $Fy$, $Fz$, $Tx$, $Ty$, $Tz$, respectively (\figref{fig:calibration}\,(c)).

\subsection{Deployment on robot}
Following \cite{chi2024universal}, the robot end-effector retains the same design as the handheld UMI-FT, including the fingers, CoinFTs, and the iPhone. The only modification is that all data modalities are streamed directly to the control desktop for real-time policy inference.

\section{Compliant Manipulation using UMI-FT}
\label{sec:methods-SW}

In this section, we introduce our software suite for compliant manipulation utilizing force/torque sensing from both fingers of UMI-FT.

\subsection{Controller architecture:}
\label{subsec:control_arch}
\begin{figure}[tb!]
\centering
	\vspace{1.5mm}
	\includegraphics[width=\linewidth]{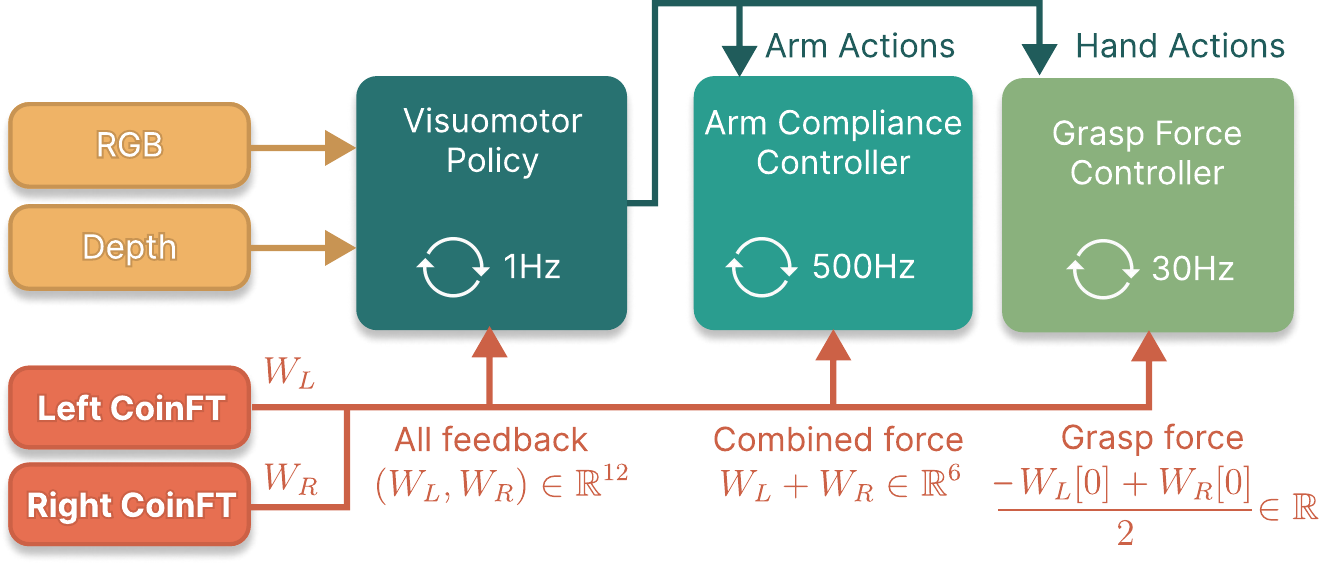}
	\caption{Controller structure of UMI-FT and the flow of information. Sensor modalities are listed on the left, with proprioception omitted for clarity. They flow to three controller loops in the middle, where the learned policy runs the slowest and generates reference targets to the other two model-based controllers.}
	\label{fig:control_structure}
	\vspace{-7pt}
\end{figure}

The force measurements from the fingers are used in three parallel loops, as illustrated in Fig.~\ref{fig:control_structure}.
The slowest loop is a learned visuomotor policy that performs manipulation reasoning. The policy takes as input the most recent 32 frames of both force sensor readings. The policy is trained per task and is detailed in Sec.~\ref{subsec:acp}. The policy outputs reference position, grasp force and stiffness, which are taken as input by the two model-based compliance controllers.

\textbf{Wrist compliance control:}
The wrist compliance controller implements 6D task space admittance control to move the robot arm like a virtual spring-mass-damper system~\cite{villani2016force}. The 6D wrench measurements from both FT sensors, $W_{S1}, W_{S2}$ are converted to the robot tool frame and combined as the wrench feedback for the admittance control, so the robot can respond to external forces on either finger or both:

\begin{equation}
    W_{wrist} = {\rm Ad}^T_{S_1T}W_{S1} + {\rm Ad}^T_{S_2T}W_{S2}
\end{equation}
where ${\rm Ad}_{S_1T}, {\rm Ad}_{S_2T}$ denotes the adjoint transformation matrix from one of the sensor frames to the robot tool frame.

Unlike in~\cite{hou2025adaptivecompliancepolicylearning} where the robot tool frame (TCP) was set close to the robot flange for comfortable kinesthetic teaching, with UMI-FT we found it helpful to set TCP to the center of the two fingertips. This allows useful and robust compliant behaviors in our manipulation tasks, such as better alignment with a surface while wiping.

\textbf{Grasp force control:}
Grasp force $f_G$ is measured as the average between the CoinFT readings of the two fingers in the grasp axis direction. Then we can regulate grasp force to a target $f_G^*$ with a velocity-resolved admittance controller:
\begin{equation}
    v_G = K_p(x_G^* - x_G) + K_f(f_G^*-f_G)
\end{equation}
where the position gain $K_p$ and force gain $K_f$ can be tuned to adjust how sensitive the grasping motion is to position and force tracking error. Grasping velocity $v_G$ is sent to the gripper controller.
Both the wrist and gripper compliance controller runs at the rate limit for their corresponding hardware (500Hz for the robot, 30Hz for the gripper).

\subsection{Imitation Learning with Compliance}
\label{subsec:acp}
\begin{figure}[tb!]
\centering
	\vspace{1.5mm}
	\includegraphics[width=\linewidth]{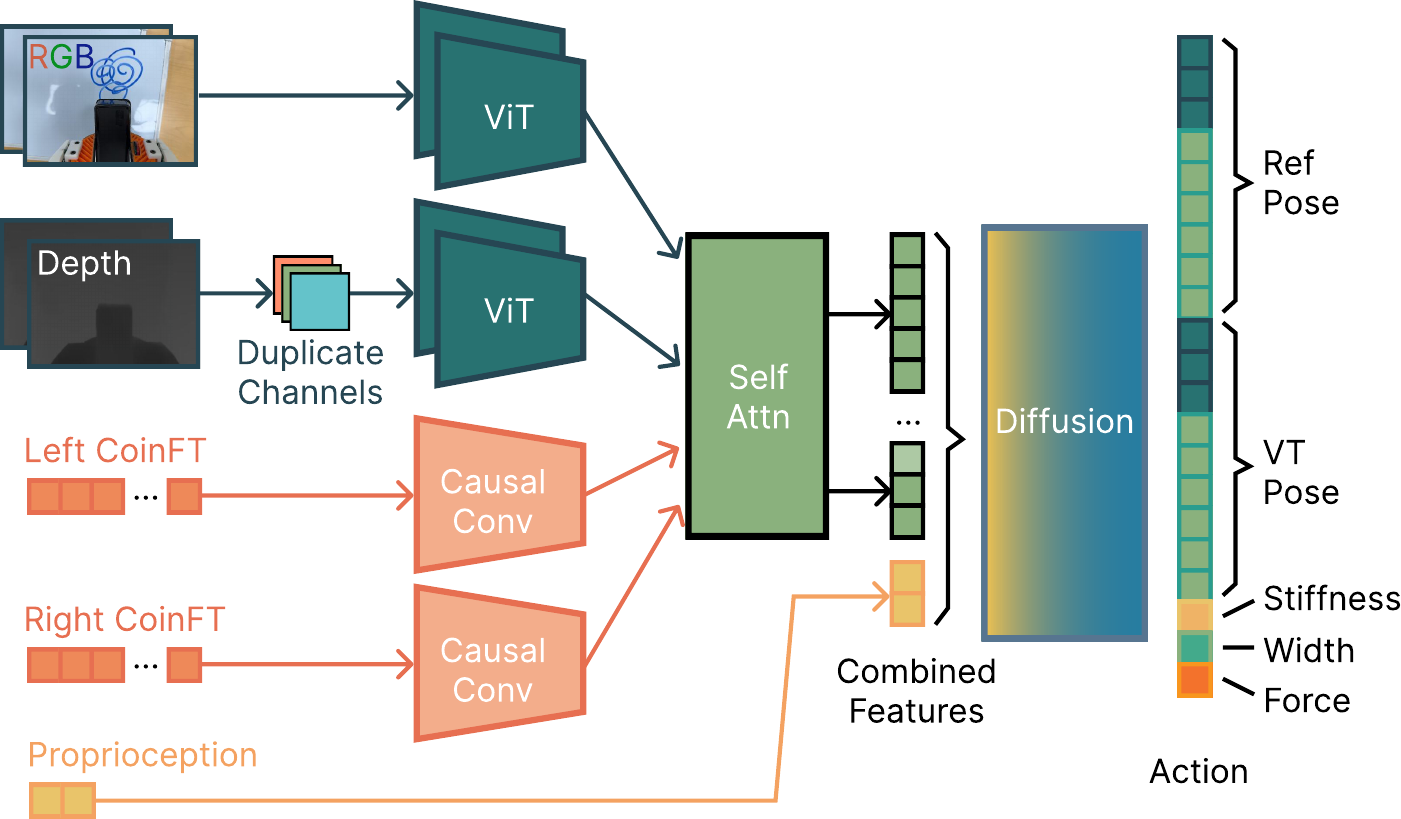}
	\caption{Structure of the adaptive compliance policy with modified inputs and outputs.}
	\label{fig:policy}
	\vspace{-7pt}
\end{figure}
We employ Adaptive Compliance Policy (ACP) \cite{hou2025adaptivecompliancepolicylearning} to learn compliant manipulation behaviors. On the high level, ACP predicts not only target positions but also target forces and desired compliance parameters, which can be executed on a robot by a standard compliance controller. With UMI-FT, we augment both inputs and outputs of ACP to accommodate hardware improvements.

\textbf{Observation Encoding:}
Our policy architecture processes multimodal observations through specialized encoders for each modality.

We sample iPhone RGB images from the two previous timesteps, with each image processed by a CLIP-pretrained ViT-B/32 encoder \cite{dosovitskiy2020image}. Images are resized to 224×224 pixels and augmented using random cropping and color jittering.

Depth input from the iPhone is processed using the same vision encoder architecture as RGB. To ensure compatibility with the RGB-based ViT model, depth images are copied across three channels, emulating a grayscale RGB image. Depth values are clipped at 0.5\,m to emphasize near-surface geometry and suppress distant background information.

Force/torque measurements from each CoinFT are encoded using a causal convolutional network~\cite{lee2019making}. The encoder takes in the previous 32 timesteps of wrench data and outputs a single feature vector per sensor.

The resulting tokens from the RGB, depth, and F/T encoders are passed to a transformer encoder layer with self-attention to obtain a combined visual-force representation.
Finally, the output of this fusion layer is concatenated with low-dimensional proprioceptive data, including end-effector pose from the previous two timesteps. The combined feature vector conditions the downstream diffusion policy~\cite{chi2024diffusionpolicy}.

\textbf{Output Decoding:}
The policy outputs the position target for the robot, the stiffness matrix, the reference grasp force, and the gripper action in a 21-dimensional vector:
\begin{itemize}[leftmargin=3mm] 
    \item Reference pose: 9D pose vector following convention in \cite{chi2024diffusionpolicy}, where the last six elements are the top two rows of a rotation matrix.
    \item Virtual target pose: another 9D pose vector representing the actual set target of the compliance controller.
    \item Stiffness value: following~\cite{hou2025adaptivecompliancepolicylearning}, this scalar value encodes the stiffness matrix of the robot arm when used together with the reference/virtual target poses.
    \item Gripper width and grasp force: two scalar values representing the desired gripper width and grasp force.
\end{itemize}
Before training, we first follow ACP and post-process the force-motion data to obtain stiffness and virtual target labels following a time-varying 3D mechanical spring. We also extract grasp forces from the CoinFT sensors.

At inference time, we first reconstruct the full stiffness matrix following ACP, then send both the stiffness matrix and the virtual target to the low level compliance controller, send gripper width and grasp force to the gripper force controller.


\section{Evaluations}
\label{sec:result}

\begin{figure*}[bt!]
\centering
	\vspace{1.5mm}
	\includegraphics[width=\textwidth]{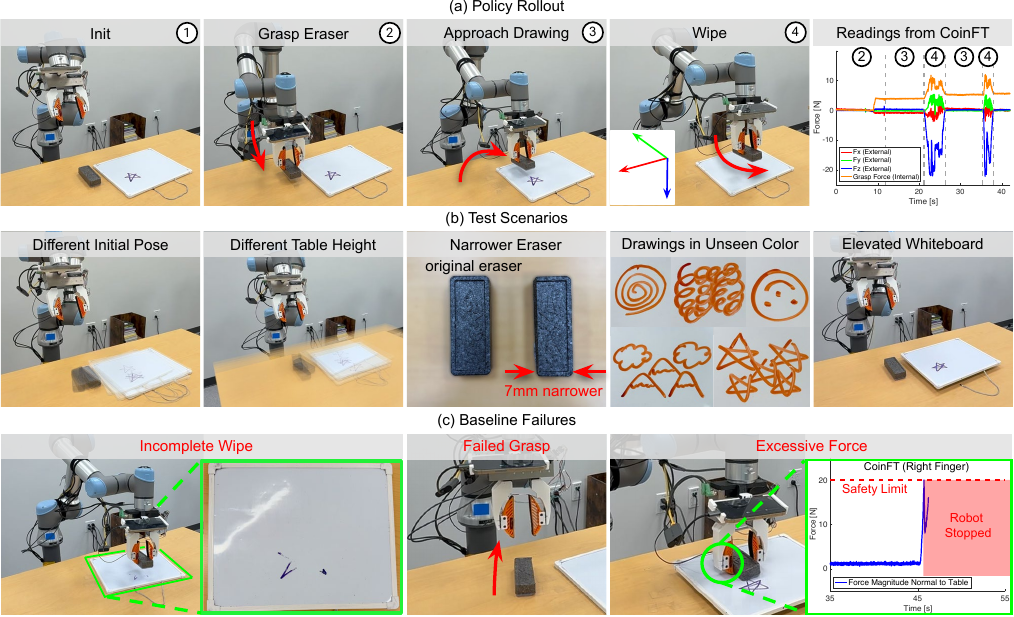}
	\caption{\textbf{Whiteboard Wiping} policy rollout, test scenarios, and representative baseline failures.}
    

	\label{fig:whiteboard}
	\vspace{-7pt}
\end{figure*}

We evaluate our approach on three tasks that highlight the benefits of compliance control with grasp force modulation. Experiments are conducted on a UR5e robot equipped with a WSG50 gripper and UMI-FT fingers. We compare our method against three baseline policies:

\begin{itemize}[leftmargin=4mm]
    \item \textbf{Diffusion Policy with Force (DP w/ F)}: This baseline takes wrench information from the CoinFT sensors as input but does not perform compliance control. The wrench signals are also used for grasp force modulation. 
    \item \textbf{Diffusion Policy (DP)}: The original diffusion policy \cite{chi2024diffusionpolicy} without force observation. The gripper is commanded purely through position control.
    \item \textbf{Diffusion Policy with Contact Microphone (DP w/ CM)}: This baseline combines diffusion policy with dynamic contact events detected via a contact microphone, following \cite{liu2024maniwav}. The contact microphone is connected to the iPhone via a USB-C mic adapter and captures audio at 44.1\,kHz. The mic is embedded on the right finger of the UMI-FT (\figref{fig:whiteboard})

\end{itemize}

For all baselines, both RGB and depth images are used as visual input. Consistent with \cite{hou2025adaptivecompliancepolicylearning}, compliance is applied in the 3-D translational space, though the framework naturally extends to full 6-D compliance when necessary.

\subsection{Whiteboard Wiping}

In this task, the robot must approach and grasp the eraser, retract, move to the drawing on the whiteboard, and wipe until the surface is clean. During grasping, the robot must apply sufficient force to firmly secure the eraser, beyond mere contact. During wiping, it must apply enough normal force to ensure effective wipe, while avoiding unnecessarily high forces that could damage the environment or the robot.

We collected 275 demonstrations for this task, incorporating variations in the initial pose, table height, and drawing patterns. Regardless of whether wrench information was included in the policy observation, we enforced safety thresholds of 25 N in the grasp direction (normal to the CoinFT sensor) and 20 N in all other directions. Task success is defined as less than a total of 1\,cm\,$\times$\,1\,cm drawing remaining in 5 wipes.

\textbf{Test Scenarios:} We applied variations not present in the training set during policy rollout. Each of the following test case was executed five times, yielding 25 rollouts per policy.

\begin{itemize}[leftmargin=4mm] 
    \item \textit{Normal}: Variations in the initial positions of the whiteboard, eraser, and location of drawing, similar to training data.
    \item \textit{Different table height}: includes values outside those seen in the training data.
    \item \textit{Unseen drawings in a new color}: Drawings were presented in an unseen color, with novel patterns.
    \item \textit{Elevated whiteboard}: The whiteboard was mounted at a higher elevation relative to the table, a configuration not seen during training.
    \item \textit{Narrower eraser}: The eraser width was reduced by 7\,mm, creating a shape not observed in training.
\end{itemize}

\begin{table}[t!]
\centering
\caption{Success Rates for Whiteboard Wiping Task.}
\begin{tabularx}{\columnwidth}{l *{6}{>{\centering\arraybackslash}X}}
\toprule
 & Normal & Board height & Table height & New color & Narrow eraser & Overall \\
\midrule
\textbf{Ours (ACP)} & 5/5 & 4/5 & 4/5 & 5/5 & 5/5 & 92\,\% \\
\textbf{DP w/ F} & 2/5 & 1/5 & 1/5 & 1/5 & 2/5 & 28\,\% \\
\textbf{DP} & 2/5 & 0/5 & 1/5 & 1/5 & 0/5 & 16\,\% \\
\textbf{DP w/ CM} & 0/5 & 0/5 & 0/5 & 0/5 & 0/5 & 0\,\% \\
\bottomrule
\end{tabularx}
\label{tab:wbw-eval}
\end{table}

\textbf{Findings:} The results of the whiteboard wiping task are summarized in \tableref{tab:wbw-eval}. Overall, high-frequency contact modulation through compliance control and grasp force regulation enables our method to reliably execute forceful tasks such as wiping, while generalizing to diverse environmental disturbances. 

\ul{A.1) Grasp force modulation improves robustness to unseen object shapes.} Our method performed reliably in the \textit{narrower eraser} scenario. The [Diffusion Policy with Force] succeeded in grasping the eraser, but failed due to incomplete wipes or excessive force for this scenario. In contrast, all baselines without force information never succeeded grasping in this scenario, and also failed intermittently in other scenarios.

\ul{A.2) Dynamic contact information alone is insufficient for continuous contact modulation.} The [Diffusion Policy with Contact Mic] baseline never succeeded, with the dominant failure mode being excessive applied force. Additional failures included grasping errors, such as failing to establish contact with the narrow eraser or barely touching the eraser without lifting it. While ManiWAV \cite{liu2024maniwav} showed that audio from a contact microphone can improve contact detection, it remains inadequate for regulating continuous force.

\subsection{Skewering Zucchini}
\subsubsection{\rel{In-Lab Experiments}}

In this task, the robot must skewer a pre-grasped zucchini slice onto a spike. The robot should approach the spike at an appropriate relative pose while maintaining a firm grasp, ensuring that the zucchini is properly punctured without falling off or excessively rotating due to slippage. Task success is defined as achieving a complete puncture with less than a 45\degree of rotation.

We collected 200 demonstrations for this task, with variations in initial relative pose of the spike, location along the finger where the zucchini is pre-grasped. The zucchini diameter had natural variations, while the slice thickness was controlled to 1\,cm. As in the \textit{whiteboard wiping} task, safety thresholds of 25\,N in the grasp direction and 20\,N in all other directions were enforced, measured by the CoinFTs.

\textbf{Test Scenarios:} Each policy was evaluated over 20 rollouts, with 5 trials per test case, to assess generalization and robustness. During rollout, the zucchini was always aligned with the center of the CoinFTs. The [Diffusion policy with contact mic] baseline was removed for this task as the gripper kept dropping the zucchini slice while approaching and never succeeded.

\begin{itemize}[leftmargin=4mm]
    \item \textit{Normal}: Variations in the initial pose of the spike, using a 1\,cm green zucchini slice, similar to training.

    \item \textit{Yellow Zucchini}: A 1\,cm slice of yellow zucchini, unseen in training.

    \item \textit{Thicker Zucchini}: A 2\,cm slice of green zucchini, unseen in training, requiring a larger puncture force.

    \item \textit{Fork}: A fork was used instead of a spike, introducing both visual variation and higher puncture force. This case was not present in the training set. 
    
\end{itemize}

\begin{figure}[tp!]
\centering
	\vspace{1.5mm}
	\includegraphics[width=\linewidth]{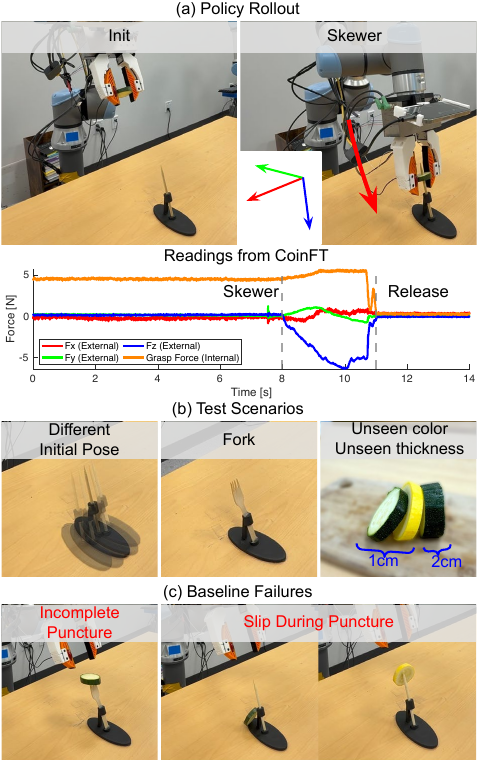}
	\caption{\textbf{Skewering Zucchini} policy rollout, test scenarios, representative baseline failures.}

	\label{fig:task-zucchini}
	\vspace{-7pt}
\end{figure}

\begin{table}[tb!]
\centering
\caption{Success Rates for Zucchini Skewering Task.}
\begin{tabularx}{\columnwidth}{l *{5}{>{\centering\arraybackslash}X}}
\toprule
 & Normal & Yellow & Thicker & Fork & Overall \\
\midrule
\textbf{Ours (ACP)} & 5/5 & 5/5 & 3/5 & 3/5 & 80\,\% \\
\textbf{DP w/ F} & 5/5 & 4/5 & 3/5 & 2/5 & 70\,\% \\
\textbf{DP} & 2/5 & 2/5 & 2/5 & 0/5 & 30\,\% \\
\bottomrule
\end{tabularx}
\label{tab:zucchini-eval}
\end{table}




    

\textbf{Findings:} The evaluation results are summarized in \tableref{tab:zucchini-eval}. We find that \ul{B.1) grasp force regulation is critical for maintaining a secure hold on the zucchini during skewering}. As shown in the [Diffusion Policy] baseline, the success rate of the task drops significantly without force information. This occurs because the gripper is controlled solely through predicted width and is unaware of contact forces, causing the zucchini to frequently slip while skewering. As shown in \figref{fig:task-zucchini}\,(c), the failures are due to linear or rotational slip. In contrast, once the grasp force regulation is included, these failures are largely eliminated. Moreover, the difference between having or not having compliance control becomes small. We hypothesize that this is because the zucchini itself is compliant, reducing the additional benefit of compliance at the manipulator level. 

\subsubsection{\rel{In-the-Wild Experiments}}

\rel{To investigate in-the-wild generalization, we collected 630 demonstrations across 15 different scenes. Other than the scene variance, the data collection protocol remained consistent with the in-lab case.}

\begin{figure}[tp!]
\centering
	\vspace{1.5mm}
	\includegraphics[width=\linewidth]{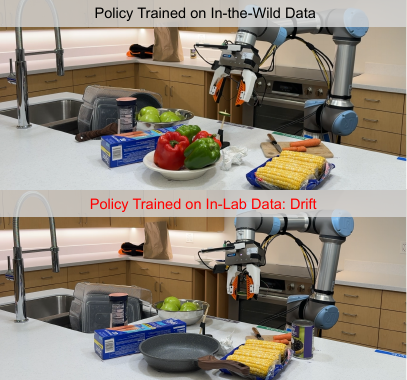}
	\caption{\rel{\textbf{Skewering Zucchini} in-the-wild.}}

	\label{fig:task-zucchini-wild}
	\vspace{-7pt}
\end{figure}

\rel{\textbf{Test Scenarios:} We compared our modified ACP policy trained on either in-lab data or in-the-wild data. These policies were evaluated over 20 rollouts, in a scene not included in either dataset with added clutter using unseen objects (\figref{fig:task-zucchini-wild}). Each rollout varied in initial pose of the spike and different sets of unseen objects were used every five rollouts.}

\rel{\textbf{Findings: } We find that \ul{B.2) in-the-wild data enables generalization to unseen environments} (\figref{fig:task-zucchini-wild}). As shown in \tableref{tab:zucchini-wild}, the policy trained on diverse data achieves a notably higher success rate (20/20), successfully navigating to the skewer while maintaining grasp force despite unseen surroundings and clutter. In contrast, the policy trained on data with limited scene-diversity rarely succeeds (4/20), where the dominant failure mode was missing the skewer due to poor navigation.}

\begin{table}[tb!]
\centering
\caption{\rel{Results for In-the-Wild Zucchini Skewering.}}
\begin{tabularx}{\columnwidth}{l *{2}{>{\centering\arraybackslash}X}}
\toprule
Method & Success Rate \\
\midrule
\textbf{ACP w/ in-the-wild data} & 20/20 \\
\textbf{ACP w/ in-lab data} & 4/20 \\
\bottomrule
\end{tabularx}
\label{tab:zucchini-wild}
\end{table}

\begin{figure*}[bt!]
\centering
	\vspace{1.5mm}\includegraphics[width=\textwidth]{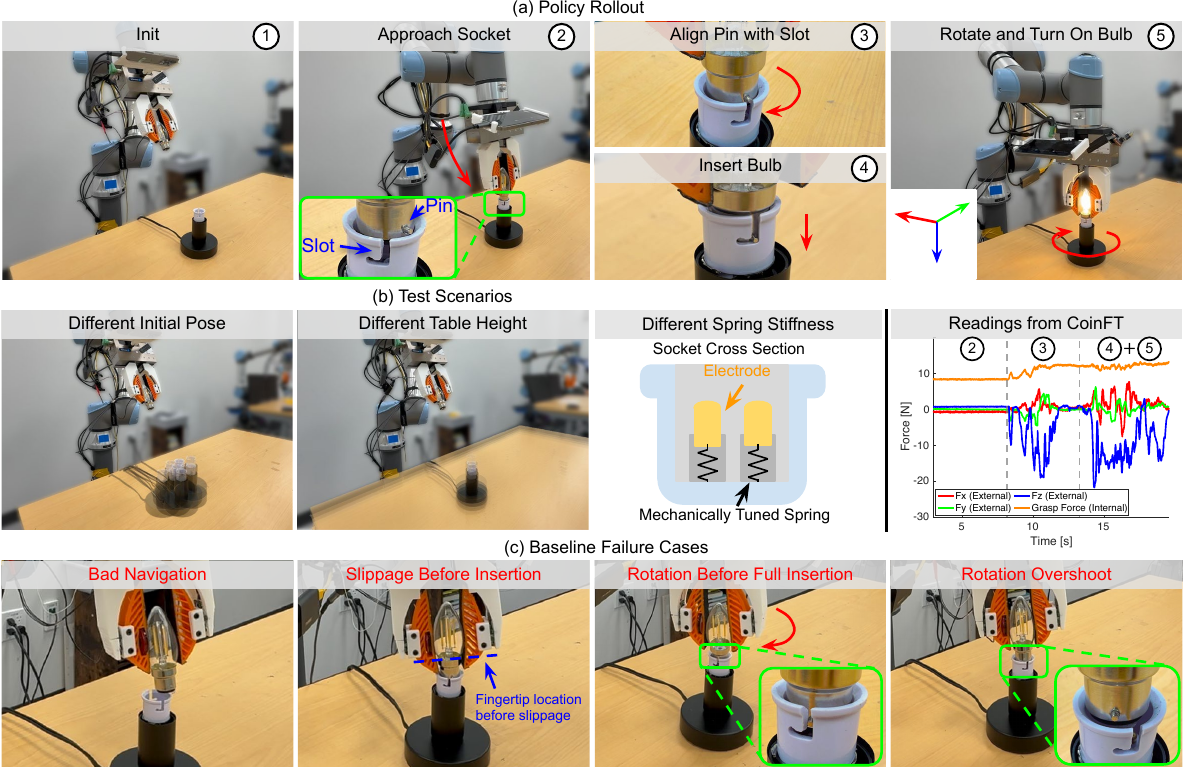}
	\caption{\textbf{Lightbulb Insertion} policy rollout, test scenarios, and representative baseline failures.}
    
	\label{fig:task-bulb}
	\vspace{-7pt}
\end{figure*}

\subsection{Lightbulb Insertion}

In this task, the robot begins with the bulb pre-grasped and must navigate toward the socket, align the bayonet pins with the slot on the socket, insert the bulb, apply sufficient force (at least 15\,N, measured using a Mark-10 digital force gauge) to overcome the spring-loaded electrode, and then rotate to turn on the light (\figref{fig:task-bulb}\,(a),\,(b)). For successful alignment and insertion, it is desirable that the bayonet pin remains in contact with the socket rim during rotation, ensuring that it drops into the slot once properly aligned. The gripper must hold the bulb firmly and apply enough force to compress the spring-loaded electrode, while avoiding excessive force that could lead to slippage. We manually shortened the spring to reduce the socket’s reaction force from 30\,N to 15\,N, as 30\,N was excessive for our setup. Task success is defined as the bulb illuminating.

We collected 200 demonstrations for this task, varying the initial socket pose, table height, and the bulb’s pre-grasp location relative to the fingers. Similar to the previous tasks, safety thresholds of 50\,N in the grasp direction and 20\,N in all other directions were enforced, measured using the CoinFT sensors.

\textbf{Test Scenarios:} Each policy was evaluated over 20 rollouts, with 10 trials for the \textit{normal} case, and 5 trials for the others, to investigate generalization and robustness.

\begin{itemize}[leftmargin=3mm]
    \item \textit{Normal:} Variations in the initial pose of the socket, similar to training.  
    \item \textit{Table Height:} Variations in table height, including values not present in the training set.  
    \item \textit{Stiffer Socket:} The socket spring was tuned to exert a higher reaction force during insertion (20\,N instead of 15\,N). This condition was not included in training.  
\end{itemize}



\begin{table}[tb!]
\centering
\caption{Success Rates for Lightbulb Insertion Task.}
\begin{tabularx}{\columnwidth}{l *{5}{>{\centering\arraybackslash}X}}
\toprule
 & Normal & Table height & Stiffer socket & Overall \\
\midrule
\textbf{Ours (ACP)} & 10/10 & 5/5 & 4/5 & 95\,\% \\
\textbf{DP w/ F} & 4/10 & 5/5 & 3/5 & 60\,\% \\
\textbf{DP} & 0/10 & 0/5 & 0/5 & 0\,\% \\
\textbf{DP w/ CM} & 3/10 & 1/5 & 0/5 & 20\,\% \\
\bottomrule
\end{tabularx}
\label{tab:bulb-eval}
\end{table}

\textbf{Findings:} The evaluation results are summarized in \tableref{tab:bulb-eval}.
\ul{C.1) Compliance control is critical for haptic search.} A crucial step in this task is to locate the slot, align the bayonet pin, and insert it. Due to occlusion from the bulb and the small size of the slot, this is difficult to achieve relying only on vision. Our method consistently succeeded by maintaining controlled contact force with the socket while rotating the bulb, allowing the pin to slide into the slot once aligned. In contrast, the baselines often failed to maintain contact during rotation, overshooting and missing the slot (\figref{fig:task-bulb}\,(c)). Even when insertion succeeded, excessive contact forces frequently caused slippage during the haptic search, sometimes resulting in rotation before full insertion. Moreover, the [Diffusion Policy] baseline showed larger navigation errors when approaching the socket compared to multimodal baselines, consistent with observations in \cite{liu2024maniwav}.


\section{Future Work and Conclusion}

The current UMI-FT data collection setup is tethered, as the USB–serial bridge connecting the two CoinFTs to the laptop relies on a USB cable. UMI-FT could be made wireless by using a Bluetooth-capable microcontroller, allowing the iPhone to directly connect to the CoinFTs. We leave this extension as future work.

Although the iPhone provides an ultrawide RGB (120\degree  \,FoV) stream, in this project we only used the main RGB camera (approx. 80\degree \,diagonal FoV). Our chosen tasks did not require a wider field of view, but the same architecture presented here can be applied in the future when larger spatial coverage is needed.

The CoinFT itself also has room for improvement. As discussed \secref{sec:methods-HW}, CoinFT can delaminate under sufficiently large tensile forces. Further design improvements such as modifying the pillar geometry or improving the bonding method could enhance mechanical reliability. In addition, the current microcontroller (PSoC 4100S) could be upgraded to a newer generation chip to improve overall performance.

In this paper, we presented UMI-FT, a scalable multimodal data collection platform equipped with custom F/T sensors on each finger. Through real-world robot experiments, we demonstrated that policies learned with UMI-FT enable compliant behaviors, allowing robots to reliably perform tasks that require careful modulation of both external contact forces and internal grasp forces.

\section{Acknowledgments}
This work was supported in part by the  NSF Award \#2143601, \#2037101, and \#2132519, Toyota Research Institute, Samsung and Amazon. We thank Google and TRI for the UR5 robot hardware. We thank Huy Ha, Zeyi Liu, and other members of the Robotics and Embodied Artificial Intelligence Lab for the fruitful discussions. We also thank Alice Wu, Eric Cousineau, Rick Cory, Jeannette Bohg for their valuable advice and insights. The views and conclusions contained herein are those of the authors and should not be interpreted as necessarily representing the official policies, either expressed or implied, of the sponsors.

\bibliographystyle{IEEEtran}
\bibliography{References}

\begin{thebibliography}{10}
\providecommand{\url}[1]{#1}
\csname url@rmstyle\endcsname
\providecommand{\newblock}{\relax}
\providecommand{\bibinfo}[2]{#2}
\providecommand\BIBentrySTDinterwordspacing{\spaceskip=0pt\relax}
\providecommand\BIBentryALTinterwordstretchfactor{4}
\providecommand\BIBentryALTinterwordspacing{\spaceskip=\fontdimen2\font plus
\BIBentryALTinterwordstretchfactor\fontdimen3\font minus \fontdimen4\font\relax}
\providecommand\BIBforeignlanguage[2]{{%
\expandafter\ifx\csname l@#1\endcsname\relax
\typeout{** WARNING: IEEEtran.bst: No hyphenation pattern has been}%
\typeout{** loaded for the language `#1'. Using the pattern for}%
\typeout{** the default language instead.}%
\else
\language=\csname l@#1\endcsname
\fi
#2}}

\bibitem{chi2024universal}
C.~Chi, Z.~Xu, C.~Pan, E.~Cousineau, B.~Burchfiel, S.~Feng, R.~Tedrake, and S.~Song, ``Universal manipulation interface: In-the-wild robot teaching without in-the-wild robots,'' in \emph{Proceedings of Robotics: Science and Systems}, 2024.

\bibitem{bhirangi2024anyskinplugandplayskinsensing}
R.~Bhirangi, V.~Pattabiraman, E.~Erciyes, Y.~Cao, T.~Hellebrekers, and L.~Pinto, ``Anyskin: Plug-and-play skin sensing for robotic touch,'' \emph{arXiv preprint arXiv:2409.08276}, 2024.

\bibitem{liu2024maniwav}
Z.~Liu, C.~Chi, E.~Cousineau, N.~Kuppuswamy, B.~Burchfiel, and S.~Song, ``Maniwav: Learning robot manipulation from in-the-wild audio-visual data,'' \emph{arXiv preprint arXiv:2406.19464}, 2024.

\bibitem{zhu2025touch}
X.~Zhu, B.~Huang, and Y.~Li, ``Touch in the wild: Learning fine-grained manipulation with a portable visuo-tactile gripper,'' \emph{arXiv preprint arXiv:2507.15062}, 2025.

\bibitem{choi2025coinftcoinsizedcapacitive6axis}
H.~Choi, J.~E. Low, T.~M. Huh, G.~A. Uribe, S.~Hong, K.~A.~W. Hoffman, J.~Di, T.~G. Chen, A.~A. Stanley, and M.~R. Cutkosky, ``Coinft: A coin-sized, capacitive 6-axis force torque sensor for robotic applications,'' \emph{arXiv preprint arXiv:2503.19225}, 2025.

\bibitem{galbally2022elly}
E.~Galbally, A.~Piedra, C.~Brosque, and O.~Khatib, ``Elly: A real-time failure recovery and data collection system for robotic manipulation,'' \emph{arXiv preprint arXiv:2208.11845}, 2022.

\bibitem{liu2025forcemimic}
W.~Liu, J.~Wang, Y.~Wang, W.~Wang, and C.~Lu, ``Forcemimic: Force-centric imitation learning with force-motion capture system for contact-rich manipulation,'' in \emph{IEEE International Conference on Robotics and Automation}, 2025.

\bibitem{hou2025adaptivecompliancepolicylearning}
Y.~Hou, Z.~Liu, C.~Chi, E.~Cousineau, N.~Kuppuswamy, S.~Feng, B.~Burchfiel, and S.~Song, ``Adaptive compliance policy: Learning approximate compliance for diffusion guided control,'' \emph{arXiv preprint arXiv:2410.09309}, 2025.

\bibitem{wu2025robocopilot}
P.~Wu, Y.~Shentu, Q.~Liao, D.~Jin, M.~Guo, K.~Sreenath, X.~Lin, and P.~Abbeel, ``Robocopilot: Human-in-the-loop interactive imitation learning for robot manipulation,'' \emph{arXiv preprint arXiv:2503.07771}, 2025.

\bibitem{li2020review}
Q.~Li, O.~Kroemer, Z.~Su, F.~F. Veiga, M.~Kaboli, and H.~J. Ritter, ``A review of tactile information: Perception and action through touch,'' \emph{IEEE Transactions on Robotics}, vol.~36, no.~6, pp. 1619--1634, 2020.

\bibitem{luo2025tactile}
S.~Luo, N.~F. Lepora, W.~Yuan, K.~Althoefer, G.~Cheng, and R.~Dahiya, ``Tactile robotics: An outlook,'' \emph{arXiv preprint arXiv:2508.11261}, 2025.

\bibitem{li2024vision}
S.~Li, Z.~Wang, C.~Wu, X.~Li, S.~Luo, B.~Fang, F.~Sun, X.-P. Zhang, and W.~Ding, ``When vision meets touch: A contemporary review for visuotactile sensors from the signal processing perspective,'' \emph{IEEE Journal of Selected Topics in Signal Processing}, vol.~18, no.~3, pp. 267--287, 2024.

\bibitem{yuan2017gelsight}
W.~Yuan, S.~Dong, and E.~H. Adelson, ``Gelsight: High-resolution robot tactile sensors for estimating geometry and force,'' \emph{Sensors}, vol.~17, no.~12, p. 2762, 2017.

\bibitem{huang20243d}
B.~Huang, Y.~Wang, X.~Yang, Y.~Luo, and Y.~Li, ``3d-vitac: Learning fine-grained manipulation with visuo-tactile sensing,'' \emph{arXiv preprint arXiv:2410.24091}, 2024.

\bibitem{zhao2025polytouch}
J.~Zhao, N.~Kuppuswamy, S.~Feng, B.~Burchfiel, and E.~Adelson, ``Polytouch: A robust multi-modal tactile sensor for contact-rich manipulation using tactile-diffusion policies,'' \emph{arXiv preprint arXiv:2504.19341}, 2025.

\bibitem{xu2025dexumiusinghumanhand}
M.~Xu, H.~Zhang, Y.~Hou, Z.~Xu, L.~Fan, M.~Veloso, and S.~Song, ``Dexumi: Using human hand as the universal manipulation interface for dexterous manipulation,'' \emph{arXiv preprint arXiv:2505.21864}, 2025.

\bibitem{DexForce_Chen2025}
C.~Chen, Z.~Yu, H.~Choi, M.~Cutkosky, and J.~Bohg, ``Dexforce: Extracting force-informed actions from kinesthetic demonstrations for dexterous manipulation,'' \emph{IEEE Robotics and Automation Letters}, vol.~10, no.~6, pp. 6416--6423, 2025.

\bibitem{li2022see}
H.~Li, Y.~Zhang, J.~Zhu, S.~Wang, M.~A. Lee, H.~Xu, E.~Adelson, L.~Fei-Fei, R.~Gao, and J.~Wu, ``See, hear, and feel: Smart sensory fusion for robotic manipulation,'' \emph{arXiv preprint arXiv:2212.03858}, 2022.

\bibitem{liu2023enhancing}
Y.~Liu, X.~Xu, W.~Chen, H.~Yuan, H.~Wang, J.~Xu, R.~Chen, and L.~Yi, ``Enhancing generalizable 6d pose tracking of an in-hand object with tactile sensing,'' \emph{IEEE Robotics and Automation Letters}, vol.~9, no.~2, pp. 1106--1113, 2023.

\bibitem{el2024compact}
A.~El-Azizi, S.~Islam, P.~Piacenza, K.~Jiang, I.~Kymissis, and M.~Ciocarlie, ``Compact led-based displacement sensing for robot fingers,'' \emph{arXiv preprint arXiv:2410.03481}, 2024.

\bibitem{villani2016force}
L.~Villani and J.~De~Schutter, ``Force control,'' in \emph{Springer handbook of robotics}.\hskip 1em plus 0.5em minus 0.4em\relax Springer, 2016, pp. 195--220.

\bibitem{8794366}
Y.~Hou and M.~T. Mason, ``Robust execution of contact-rich motion plans by hybrid force-velocity control,'' in \emph{International Conference on Robotics and Automation}, 2019, pp. 1933--1939.

\bibitem{hou2020manipulation}
Y.~Hou, Z.~Jia, and M.~T. Mason, ``Manipulation with shared grasping,'' in \emph{Robotics: Science and Systems}, 2020.

\bibitem{lee2019making}
M.~A. Lee, Y.~Zhu, K.~Srinivasan, P.~Shah, S.~Savarese, L.~Fei-Fei, A.~Garg, and J.~Bohg, ``Making sense of vision and touch: Self-supervised learning of multimodal representations for contact-rich tasks,'' in \emph{International conference on robotics and automation}.\hskip 1em plus 0.5em minus 0.4em\relax IEEE, 2019, pp. 8943--8950.

\bibitem{luo2025precise}
J.~Luo, C.~Xu, J.~Wu, and S.~Levine, ``Precise and dexterous robotic manipulation via human-in-the-loop reinforcement learning,'' \emph{Science Robotics}, vol.~10, no. 105, 2025.

\bibitem{xu2025compliant}
X.~Xu, Y.~Hou, Z.~Liu, and S.~Song, ``Compliant residual dagger: Improving real-world contact-rich manipulation with human corrections,'' \emph{arXiv preprint arXiv:2506.16685}, 2025.

\bibitem{hogan1985impedance}
N.~Hogan, ``Impedance control: An approach to manipulation: Part ii—implementation,'' \emph{Journal of dynamic systems, measurement, and control}, vol. 107, no.~1, pp. 8--16, 1985.

\bibitem{maples1986experiments}
J.~Maples and J.~Becker, ``Experiments in force control of robotic manipulators,'' in \emph{IEEE International Conference on Robotics and Automation}, vol.~3, 1986, pp. 695--702.

\bibitem{dosovitskiy2020image}
A.~Dosovitskiy, L.~Beyer, A.~Kolesnikov, D.~Weissenborn, X.~Zhai, T.~Unterthiner, M.~Dehghani, M.~Minderer, G.~Heigold, S.~Gelly, \emph{et~al.}, ``An image is worth 16x16 words: Transformers for image recognition at scale,'' \emph{arXiv preprint arXiv:2010.11929}, 2020.

\bibitem{chi2024diffusionpolicy}
C.~Chi, Z.~Xu, S.~Feng, E.~Cousineau, Y.~Du, B.~Burchfiel, R.~Tedrake, and S.~Song, ``Diffusion policy: Visuomotor policy learning via action diffusion,'' \emph{The International Journal of Robotics Research}, 2024.

\end{thebibliography}

\end{document}